\documentclass[runningheads]{llncs}
\usepackage{graphicx}
\usepackage{amsmath,amssymb} 
\usepackage{color}
\usepackage{booktabs}
\usepackage{tabularx}
\usepackage[ruled,algonl,lined]{algorithm2e}
\usepackage{pifont}
\usepackage{hyperref}

\newcommand{\eg}{\textit{e}.\textit{g}., }

\begin{document}
\pagestyle{headings}
\mainmatter
\def\ACCV20SubNumber{***}  

\title{Visual Tracking by \\ TridentAlign and Context Embedding} 
\titlerunning{Visual Tracking by TridentAlign and Context Embedding}
\authorrunning{J. Choi et al.}

\author{Janghoon Choi$^{1}$, Junseok Kwon$^{2}$, and Kyoung Mu Lee$^{1}$}
\institute{$^{1}$ASRI, Department of ECE, Seoul National University\\
	$^{2}$School of CSE, Chung-Ang Univeristy\\
	{\tt\small ultio791@snu.ac.kr},
	{\tt\small jskwon@cau.ac.kr},
	{\tt\small kyoungmu@snu.ac.kr}
}

\maketitle

\begin{abstract}
Recent advances in Siamese network-based visual tracking methods have enabled high performance on numerous tracking benchmarks. However, extensive scale variations of the target object and distractor objects with similar categories have consistently posed challenges in visual tracking. 
To address these persisting issues, we propose novel TridentAlign and context embedding modules for Siamese network-based visual tracking methods. 
The TridentAlign module facilitates adaptability to extensive scale variations and large deformations of the target, where it pools the feature representation of the target object into multiple spatial dimensions to form a feature pyramid, which is then utilized in the region proposal stage. 
Meanwhile, context embedding module aims to discriminate the target from distractor objects by accounting for the global context information among objects. 
The context embedding module extracts and embeds the global context information of a given frame into a local feature representation such that the information can be utilized in the final classification stage. 
Experimental results obtained on multiple benchmark datasets show that the performance of the proposed tracker is comparable to that of state-of-the-art trackers, while the proposed tracker runs at real-time speed.\footnote{Code available on \textcolor{magenta}{\href{https://github.com/JanghoonChoi/TACT}{\texttt{\small https://github.com/JanghoonChoi/TACT}}}}
\end{abstract}

\section{Introduction}
\label{sec:intro}
Visual tracking, which has practical applications such as automated surveillance, robotics, and image stabilization, is one of the fundamental problems among the fields under computer vision research.
Given initial target bounding box coordinates along with the first frame of a video sequence, visual tracking algorithms aim to precisely locate the target object in the subsequent frames of the video. 
Tracking algorithms are designed to successfully track targets under various circumstances such as scale change, illumination change, occlusion, deformation, and motion blur.

Along with the wide application of convolutional neural networks (CNNs) to various computer vision tasks \cite{alexnet,resnet,frcnn}, recent advances in Siamese network-based visual tracking methods \cite{siamfc,siamrpn,siamrpn++,siamdw} have advantages in performance and speed owing to the scalability of the network and elimination of online updates. 
However, most existing Siamese trackers are still designed and trained for short-term tracking scenarios \cite{otb,vot} with strong assumptions of motion smoothness and gradual scale change. 
In order to impose these assumptions, most trackers assume a small search region around the target and penalize large displacements using hand-crafted window functions. 
This makes the tracker susceptible to error accumulation and drift due to erroneous localizations, leading to lower performance in numerous long-term tracking scenarios \cite{LaSOT,trackingnet,oxuva} in which smoothness assumptions are no longer valid. 
To alleviate these problems, recent Siamese trackers have adopted full-frame search \cite{globaltrack,siamrcnn}, so that the target location can be recovered after the out-of-frame disappearance, at the cost of computation time and with sub-real-time speeds. 
Although a full-frame search can be effective for re-detecting the target, it is susceptible to distractors with appearances similar to that of the target owing to the lack of temporal consistency and global context modeling. 
Diverse scale variations of the target can also lead to failed re-detection because the target feature representation has a fixed spatial dimension that may fail to represent the wide spatial variation of the target object.

\begin{figure}[t]
	\begin{center}
		\includegraphics[width=0.95\linewidth]{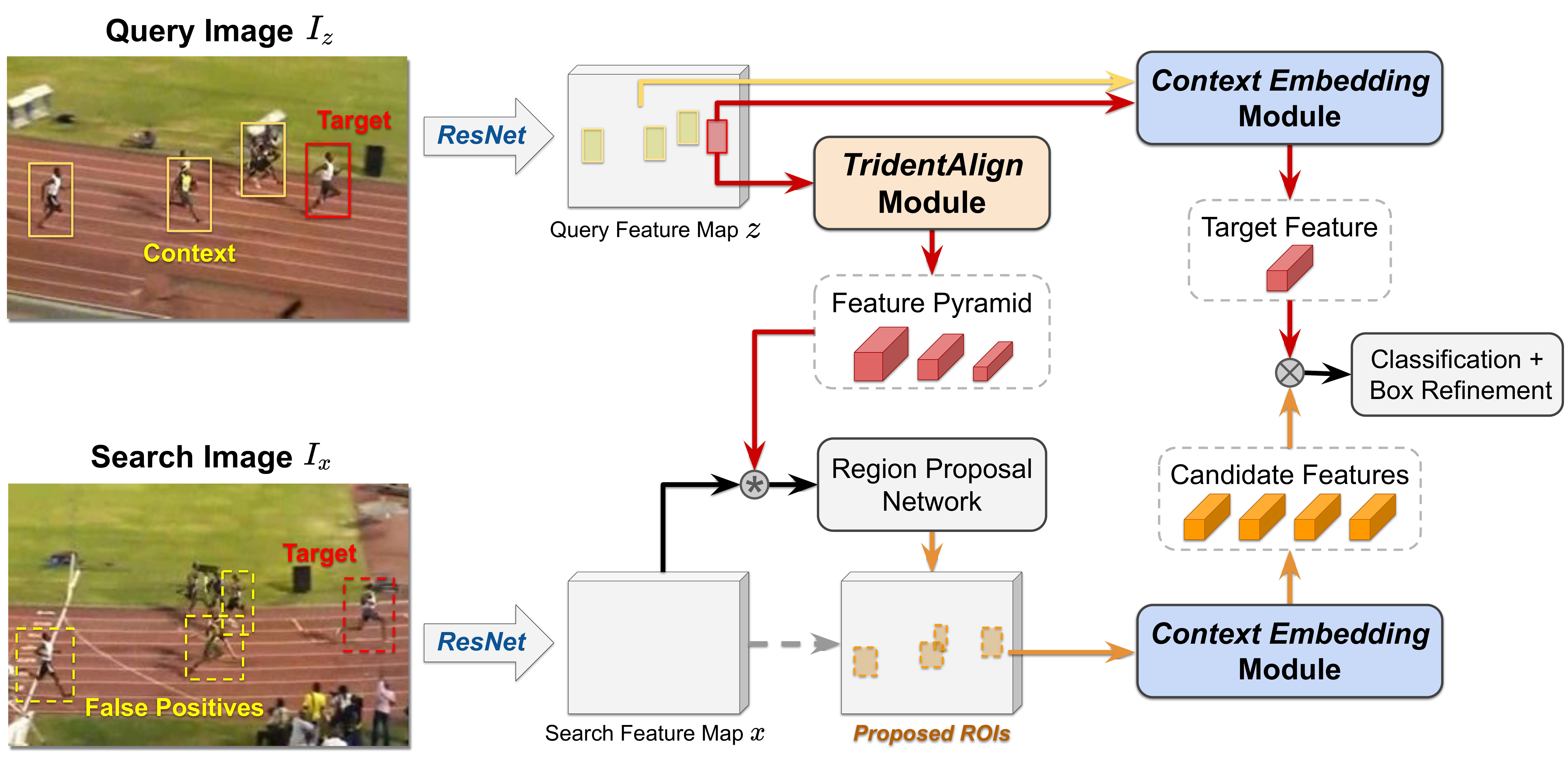}
	\end{center}
	\vspace{-7mm}
	\caption{\textbf{Overview of the proposed visual tracking framework.} Our tracker incorporates the TridentAlign module, which  generates a feature pyramid representation of the target that can be utilized for better scale adaptability of the RPN. Moreover, the context embedding module modulates the locally pooled features to incorporate the global context information of a given frame, which encourages the discrimination of the target object from false positives.}
	\label{fig:overview}
\end{figure}


In this paper, we propose a novel real-time visual tracking algorithm to address the aforementioned issues by incorporating the TridentAlign and context embedding modules into our tracking framework, where our region proposal network (RPN) is inspired by the success of recent object detectors. 
We incorporate the anchor-free FCOS \cite{fcos} detector head in our RPN to minimize the number of parameters while maximizing the flexibility with regard to large deformations in the aspect ratio. 
To further enforce the scale adaptiveness of the RPN, we use the target feature representation obtained from the TridentAlign module, which pools the features from the target region to multiple spatial dimensions to form a feature pyramid, where similar approaches have been shown to be effective in \cite{sppnet,retinanet,tridentnet,hypercolumns}. 
Moreover, rather than focusing only on the similarity between the locally obtained features, we make full use of the global context information obtained from the given frames. 
The context-embedded features obtained from the context embedding module are utilized to discriminate the target from the distractor objects in the background region to avoid choosing false-positives. 
The context embedding module receives hard negative features pooled from a given frame that encompasses potential distractors. 
It then modulates the local feature representations of the candidate regions, providing additional information on global context.


We compare our method to other Siamese network-based trackers on large-scale visual tracking datasets LaSOT \cite{LaSOT}, OxUvA \cite{oxuva}, TrackingNet \cite{trackingnet}, and GOT-10k \cite{got10k} thereby demonstrating performance comparable to that of state-of-the-art trackers. 
Moreover, our proposed modules require minimal computational overhead, and our tracker is able to run at real-time speed. Our overall tracking framework is shown in Figure \ref{fig:overview}.

\section{Related Work}
\label{sec:related}

\textbf{CNN-based trackers :~}~Conventional online visual tracking algorithms solve the tracking problem via tracking-by-detection, where they attempt to locate the target inside a search region by finding the position where the classifier produces the highest similarity/classification score to that of the target. 
Given the powerful representation capabilities of CNNs, recent trackers use CNNs for feature representation and classification. 
One early CNN-based tracker \cite{nipstrack} used the feature representation obtained from the denoising autoencoder network. 
In MDNet \cite{mdnet}, VGG \cite{vgg} features with multi-task training are used, and this tracker is accelerated to real-time speed in \cite{rtmdnet} by using ROIAlign \cite{maskrcnn}. 
Correlation filter-based trackers \cite{kcf,mosse} are also widely used on top of pretrained deep features. 
Notable approaches include the use of the continuous convolutional operator for the fusion of multi-resolution CNN features \cite{ccot,eco}, spatially regularized filters \cite{deepsrdcf}, the fusion of deep and shallow features \cite{updt}, and group feature selection~\cite{gfs}.

\noindent
\textbf{Siamese network-based trackers :~}~Siamese network-based trackers have gained attention owing to their simplicity and high performance. 
SiamFC \cite{siamfc} proposed a fully convolutional end-to-end approach to visual tracking, enabling increased speed and accuracy, in which correlation filter can be also trained on top of the feature map \cite{cfnet}. 
SiamRPN \cite{siamrpn} added RPN in the final stage for more accurate localization and size estimation, and DaSiamRPN \cite{dasiamrpn} improved the discriminability of the network by adding negative pairs during training to rule out distractors. 
Both \cite{siamrpn++} and \cite{siamdw} employed deeper and wider backbone networks based on \cite{resnet} and \cite{googlenet} with extensive experimental analysis to achieve further performance gains. 
Other notable approaches include the use of a dynamic network with a transformation learning model \cite{dsiam}, pattern detection for local structure-based prediction \cite{stsiam}, and cascaded region proposal for progressive refinement \cite{crpn}. 
Recently, noteworthy methods have been proposed to improve the discriminability of the network by performing adaptation at test-time. 
Particularly, \cite{gradnet} and \cite{mlt} used gradient information obtained during tracking to update the target representation. Moreover, \cite{updatenet} used a learned update module to obtain the updated accumulated template. 
An optimization-based model predictor is used in \cite{dimp} to predict the classifier weights.

\noindent
\textbf{Context-aware trackers :~}~Conventional visual tracking algorithms treat all candidate regions in the search image independently. 
They perform tracking by choosing the candidate region with the highest similarity score to the target, where other candidate regions have no influence on this decision process. 
Online adaptation-based approaches partially address this issue by updating the tracker with previously obtained self-labeled training examples at test-time \cite{mdnet,mlt,dimp}; however, the possibility of error accumulation and drift persists. Several tracking approaches take a broader context area into consideration \cite{traca,cacf}. 
Particularly, \cite{traca} employs a context-aware network in the initial adaptation process to choose an expert autoencoder that is used throughout the online tracking process. 
Furthermore, \cite{cacf} proposed a generic context-aware correlation filter learning scheme that can add contextual information for better background suppression. 
Nevertheless, the context modeling in these approaches is still limited to a fixed spatial region around the target, which lacks the global context information of all the possible distractors in a given scene.

\noindent
\textbf{Long-term trackers :~}~With growing interest on long-term visual tracking tasks and more benchmarks becoming available to the public \cite{LaSOT,tlp,oxuva}, tracking algorithms focused on solving long-term tracking scenarios have emerged. 
Long-term visual tracking problems are typically defined as problems involving the tracking of a target object in a relatively long (\eg $>$ 1 min) video sequence, during which the target may disappear for a brief period \cite{oxuva}. 
The main challenge of long-term tracking tasks is recovering the target after its disappearance or a tracking failure. 
To solve this problem, \cite{dasiamrpn,lct} attempted to expand the search area when the confidence value falls below a predefined threshold. 
Moreover, \cite{globaltrack,siamrcnn} performed a full-frame search without any locality assumption. 
Our approach shares some similarities with \cite{globaltrack}, but it differs significantly in these aspects: (1) We use the proposed TridentAlign module to construct a feature pyramid representation of the target for improved scale adaptability and part-based predictions in the region proposal stage, whereas \cite{globaltrack} pools the target feature into a $1\times1$ kernel, losing spatial information; (2) Our tracker introduces the context embedding module, which employs hard negative features obtained from potential distractors as contextual information, and the local feature representations can be modulated accordingly during tracking. 
In contrast, \cite{globaltrack} only uses local features without any consideration of the context, which makes it susceptible to distractors; (3) Our tracker runs at a real-time speed of $57$ fps with a ResNet-18 backbone and $42$ fps with a ResNet-50 backbone.
It thus achieves higher performance on LaSOT \cite{LaSOT} with a significantly lighter backbone network (ResNet-18) and is approximately $9$ times faster than the model proposed in \cite{globaltrack}, which uses a significantly deeper backbone network (ResNet-50).

\section{Proposed Method}
\label{sec:proposed}

Inspired by two-stage object detection networks \cite{frcnn,sppnet,retinanet}, our framework largely includes two stages: the region proposal stage and classification stage. 
Given a pair of input RGB images, that is, the query image (initial frame) $I_z$ and search image (current frame) $I_x$ along with the shared backbone network $\varphi(\cdot)$, the respective feature maps $z=\varphi(I_z)$ and $x=\varphi(I_x)$ are obtained and then passed to the RPN. 
In the region proposal stage, the RPN generates proposals for target region-of-interest (RoIs) in the search image, given the target information obtained from the query image. 
Subsequently, in the classification stage, the classifier performs binary classification on the ROI proposals obtained in the previous stage, with a positive result indicating the target region and a negative indicating the background region. Figure \ref{fig:overview} shows the overall flow of the proposed method. 

In the following subsections, we provide more detailed explanations of the proposed TridentAlign module used in the region proposal stage and context embedding module used in the classification stage. 
Then, we describe the online tracking procedure for our approach.

\subsection{Region Proposal with Scale Adaptive TridentAlign}
\label{sec:tridentalign}

The initial target bounding box coordinates and input feature maps are given as $z,x \in \mathbb{R}^{h \times w \times c}$, where the channel dimension of the input feature maps are reduced from the original outputs of the ResNet backbone by employing  $1\times1$ conv layers. 
Using these feature maps, the TridentAlign module performs multiple ROIAlign \cite{maskrcnn} operations on $z$ with varying spatial dimensions to obtain target feature representations $z_i \in \mathbb{R}^{s_i \times s_i \times c}$, where $s_i$ is the spatial dimension of the pooled features. 
These features form a feature pyramid denoted as $Z=\{z_1,z_2,...,z_K\}$, where $K$ is the total number of features in the feature pyramid. 
Then, the depth-wise cross-correlation between search feature map $x$ and each target feature $z_i$ in the feature pyramid $Z$ is calculated as
\begin{equation}
    \hat{x}_i = x \circledast z_i,
\end{equation}
where $\circledast$ denotes the depth-wise cross-correlation operator with zero padding. 
As a result, each $\hat{x}_i \in \mathbb{R}^{h \times w \times c}$ is obtained for the corresponding $z_i$ and is concatenated along the channel dimension to form the multi-scale correlation map $[\hat{x}_1, \hat{x}_2, ..., \hat{x}_K ]=\hat{x} \in \mathbb{R}^{h \times w \times Kc}$. 
The correlation map is then refined as in $f_{att}(\hat{x}) \in \mathbb{R}^{h \times w \times c}$ using a self-attention block analogous to that employed in \cite{cbam}.
In this self-attention block, the adaptive channel and spatial attention weights are applied to focus on a specific position and target scale, followed by a $1\times1$ conv layer, thereby reducing the channel dimension back to $c$.

\begin{figure}[t]
	\begin{center}
		\includegraphics[width=0.95\linewidth]{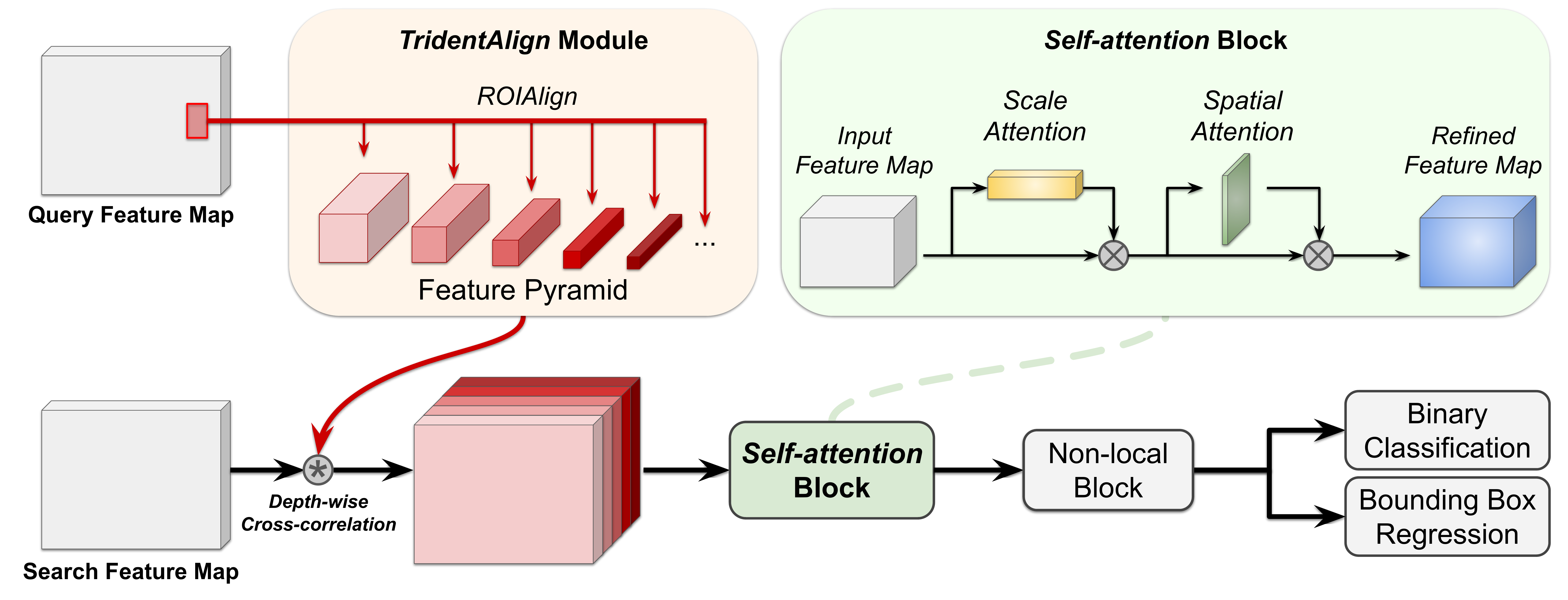}
	\end{center}
		\vspace{-7mm}
	\caption{\textbf{Overview of the proposed region proposal network.} The feature pyramid representation of the target is constructed using our TridentAlign module, wherein each feature undergoes a depth-wise cross-correlation operation with the search feature map. 
	The correlated feature maps are concatenated along the channel dimension; here, a self-attention block is used to focus more on a certain spatial area with a certain target scale. Followed by a non-local block \cite{nonlocal} and binary classification/bounding box regression branches, ROI can be obtained.}
	\label{fig:rpn}
\end{figure}

\begin{figure}[t]
	\begin{center}
		\includegraphics[width=0.95\linewidth]{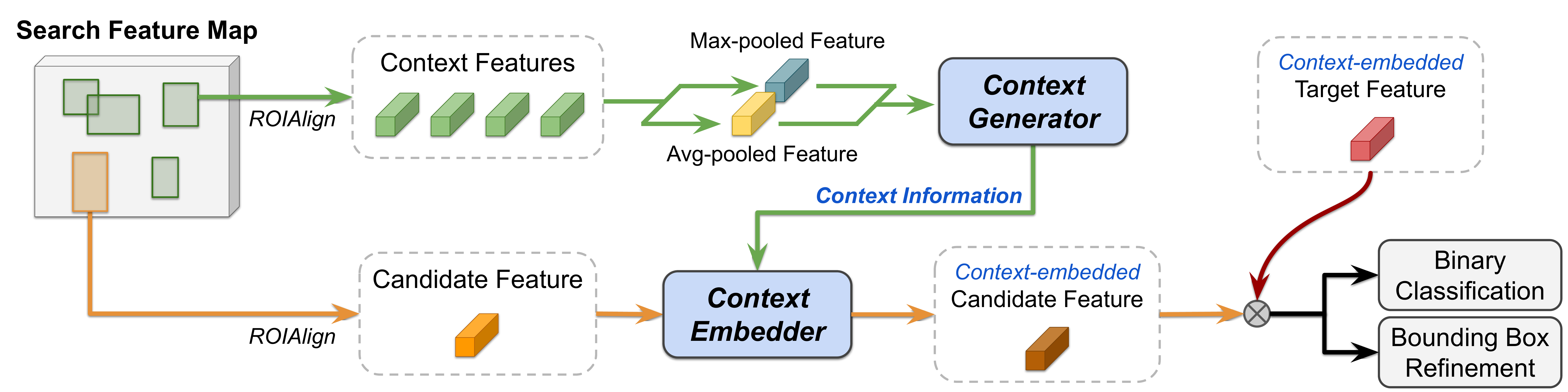}
	\end{center}
			\vspace{-7mm}
	\caption{\textbf{Overview of our context embedding framework.} Given the candidate ROI and context regions, the respective feature representations are obtained by performing ROIAlign operations on each region. 
	Using the context features, max-pooled and average-pooled features are obtained via element-wise maximum and averaging operations. 
	The context generator receives these features to generate the global context information, which the context embedder embeds into the candidate features. 
	Finally, the context-embedded candidate feature can be compared with the context-embedded target features for binary classification and bounding box refinement.}
	\label{fig:context}
\end{figure}

With the refined correlation map, we use the detection head module employed in \cite{fcos}, where each branch outputs binary classification labels and bounding box regression values. 
For a single location $(i,j)$ inside the output map, classification labels $p_{i,j}$ with bounding box regression values $t_{i,j}$ are predicted from the respective branches. 
At training stage, if a location $(i,j)$ is inside the ground-truth (GT) target bounding box, it is considered a positive sample and we assign the GT label $c^*_{i,j}=1$ and GT regression target $t^*_{i,j}=(l^*,t^*,r^*,b^*)$, where $l^*,t^*,r^*$, and $b^*$ are the distances from $(i,j)$ to the four sides (left, top, right, bottom) of the bounding box, respectively. For negative samples, we assign $c^*_{i,j}=0$.
To train the overall RPN, we use the same loss as the one used in \cite{fcos}:
\begin{equation}
    L_{rpn}(\{p_{i,j}\},\{t_{i,j}\}) = \frac{1}{N_{pos}}\sum_{i,j}{L_{cls}}{(p_{i,j},c^*_{i,j})} + \frac{\lambda}{N_{pos}}\sum_{i,j}{\mathbf{1}_{\{c^*_{i,j}>0\}}L_{reg}(t_{i,j},t^*_{i,j})},
\label{eq:loss_rpn}
\end{equation}
where $N_{pos}$ is the number of positive samples, $L_{cls}$ is the focal loss \cite{retinanet}, and $L_{reg}$ is the linear IoU loss. 
The loss is summed over all locations of the output map, and $L_{reg}$ is only calculated for positive samples. 
Subsequently, a non-maximum suppression (NMS) operation is performed to obtain the top $N$ candidate ROIs. 
The overall architecture of the RPN is illustrated in Figure \ref{fig:rpn}.

\subsection{Classification with Context-Embedded Features}
\label{sec:negfeat}

Given the candidate ROIs obtained from the preceding region proposal stage, ROIAlign operations are performed on the search feature map $x$ to obtain a set of candidate features $X=\{x_1,x_2,...,x_N\}$, where each $x_i \in \mathbb{R}^{s \times s \times c}$ with $N$ candidate regions. 
Using all of the candidate features in $X$ to generate the global context information, we aim to modulate each feature ${x}_i$ to obtain the context-embedded feature $\tilde{x}_i \in \mathbb{R}^{s \times s \times c}$. 
First, element-wise maximum and averaging operations are performed over all features in $X$ to obtain max-pooled and average-pooled features, which are concatenated along the channel dimension as $x_{cxt} \in \mathbb{R}^{s \times s \times 2c}$. 
Then, inside the context embedding module, context generator $g_{1}(\cdot)$ receives $x_{cxt}$ to generate the global context information, and context embedder $g_{2}(\cdot)$ receives both the candidate feature $x_i$ and context information from $g_1$ to generate the context-embedded feature $\tilde{x}_i$.
The overall context embedding scheme is illustrated in Figure \ref{fig:context}.
For our context generator and embedder design, we test $4$ variants: (1) simple concatenation, (2) simple addition, (3) CBAM \cite{cbam}, and (4) FILM \cite{film} based modules. 
The details of each variant are listed in Table \ref{tab:context}.

For the simple concatenation-based module (Table \ref{tab:context}(1)), $x_{cxt}$ is directly used as context information and the context embedder $g_2$ receives $[x_i,x_{cxt}]$ as input, where $[\cdot,\cdot]$ denotes concatenation along the channel dimension. In the simple addition-based module (Table \ref{tab:context}(2)), context information is generated in a form of additive modulation $\delta$, which is added to the original candidate feature $x_i$. 
For the CBAM-based module (Table \ref{tab:context}(3)), context information is generated and applied to $x_i$ as channel attention $m_c$ and spatial attention $m_s$. 
Finally, the FILM-based module (Table \ref{tab:context}(4)) modulates $x_i$ by applying an affine transformation with coefficients $\gamma$ and $\beta$. 

\begin{table}[b]
	\caption{\textbf{Variants of the context embedding module.} We test 4 possible implementations of the context embedding module. For 3-layer CNNs, we use $1\times1$ kernels with output channels set to $c$, followed by ReLU activation.}
	\label{tab:context}
	\centering
	\resizebox{\textwidth}{!}{
		\begin{tabular}{llll}
			\toprule
			\multicolumn{1}{c}{} & \multicolumn{2}{l}{Generator $g_1(x_{cxt})$} & Embedder $g_2(g_1(x_{cxt}),x_i)$ \\ \cmidrule(l){2-4} 
			& Type       & Output          & Operation \\ \cmidrule(l){1-4} 
			(1) Simple concat.~   & Identity        & $x_{cxt} \in \mathbb{R}^{s \times s \times 2c}$        & \scriptsize{3-layer CNN, $g_{2}([x_i,x_{cxt}])$} \\
			(2) Simple add.      & 3-layer CNN     & $\delta \in \mathbb{R}^{s \times s \times c}$         & $x_i + \delta$ \\
			(3) CBAM-based       & 3-layer CNN~~     & \scriptsize{$m_c \in \mathbb{R}^{1 \times 1 \times c}, m_s \in \mathbb{R}^{s \times s \times 1}$}            & $(x_i \otimes m_c) \otimes m_s$ \\
			(4) FILM-based       & 3-layer CNN     & $\gamma,\beta \in \mathbb{R}^{s\times s\times c}$   & $\gamma \otimes x_i + \beta$ \\
			\bottomrule
		\end{tabular}
	}
\end{table}

Finally, each context-embedded candidate feature $\tilde{x}_i$ is compared with the context-embedded target feature $\tilde{z}_0 \in \mathbb{R}^{s\times s\times c}$ by element-wise multiplication as in $\tilde{x}_i \otimes \tilde{z}_0$.
Binary classification and bounding box refinement operations are subsequently performed. 
For every $\tilde{x}_i$, a classification label ${c}_i$ and refined bounding box coordinates ${t}_i$ are obtained. 
At training stage, the GT classification label ${c}_i^*=1$ is assigned to candidate boxes with $\mathrm{IoU}({t}_i,t_i^*) > \tau_{p}$, where $t_i^*$ is the GT box coordinates, and $c_i^*=0$ is assigned to candidate boxes with $\mathrm{IoU}(t_i,t_i^*) < \tau_{n}$.
In our experiments, we use $(\tau_{p},\tau_{n})=(0.5,0.4)$. 
To train our context embedding module and classifier, we minimize the loss function given as
\begin{equation}
    L_{det}(\{c_i\}, \{t_i\}) = \frac{1}{N_{pos}}\sum_{i}{L_{cls}}{(c_{i},c^*_{i})} + \frac{\lambda}{N_{pos}}\sum_{i}{\mathbf{1}_{\{c^*_{i}>0\}}L_{reg}(t_{i},t^*_{i})},
\label{eq:loss_det}
\end{equation}
where the loss functions $L_{cls}$ and $L_{reg}$ are the same as those in Eq. \eqref{eq:loss_rpn}.

\begin{algorithm}[t]
	\small{
		\SetKwData{Left}{left}\SetKwData{This}{this}\SetKwData{Up}{up}
		\SetKwFunction{Union}{Union}\SetKwFunction{FindCompress}{FindCompress}
		\SetKwInOut{Input}{Input}\SetKwInOut{Output}{Output}
		
		\Input{Tracking sequence of length $L$, $\{I^1,I^2,...,I^L\}$ \\
			Initial target bounding box coordinates}
		\Output{Target bounding box coordinates for each frame}
		\BlankLine
		
		\emph{\# Initialization at $t=1$}\\
		Compute query feature map $z=\varphi(I^1)$ for initial frame $I^1$\\
		Build target feature pyramid $Z$ from $z$ using TridentAlign\\
		Using same $z$ as search feature map; obtain candidate features using RPN\\
		Obtain context-embedded target feature $\tilde{z}_0$\\
		
		\BlankLine
		\emph{\# For later frames in tracking sequence}\\
		\For{$t=2$ to $L$} {
			Compute search feature map $x=\varphi(I^t)$ for frame $I^t$\\
			Using $Z$ and $x$, obtain ROIs with candidate features $X$ using RPN\\
			For every $x_i \in X$, calculate context-embedded feature $\tilde{x}_i$ to form $\tilde{X}$\\
			Compute $\tilde{x}_i \otimes \tilde{z}_0$ for every $\tilde{x}_i \in \tilde{X}$\\
			For every ROI, obtain softmax classification scores and box refinement values\\
			Choose refined ROI with highest classification score as output\\
		}
		\caption{\small{TACT}} \label{alg:track}
	}
\end{algorithm}

\subsection{TridentAlign and Context Embedding Tracker}
\label{sec:tracking}
Herein, we propose \textbf{T}rident\textbf{A}lign and \textbf{C}ontext embedding \textbf{T}racker (\textbf{TACT}).
The overall tracking procedure is organized and shown as Algorithm \ref{alg:track}. 
The tracking process is purposely made simple to achieve real-time speed. 
Furthermore, our tracking algorithm performs a full-frame search for every frame without any motion smoothness assumption based on the previous positions of the target; therefore, it is possible to run our tracker on a batch of multiple frames of offline videos. Increasing the batch size from $1$ to $8$ results in a large boost in tracking speed: we obtain 57 $\rightarrow$ 101 fps with the ResNet-18 backbone and 42 $\rightarrow$ 65 fps with the ResNet-50 backbone.

\section{Experiments}
\label{sec:experimental}
In this section, we specify the implementation details and the experimental setup, then we compare the performance of our proposed tracking algorithm with that of other approaches on four large-scale tracking benchmarks, namely LaSOT \cite{LaSOT}, OxUvA \cite{oxuva}, TrackingNet \cite{trackingnet}, and GOT-10k \cite{got10k}. 
We also perform ablation experiments for the individual components to further analyze the effectiveness of our proposed method.

\subsection{Implementation Details}
\label{sec:implementation}
\textbf{Parameters:~}~We resized the input images to $666\times400$, where original aspect ratio is preserved by adding zero-padding to the right or bottom side of the images. 
We used ResNet-18 and ResNet-50 \cite{resnet} as the backbone feature extractor network, followed by $1\times1$ conv layers, where the channel dimension of the output features was set to $c=256$. 
We reduced the stride of the last residual block to $1$ to obtain feature maps with a size of $42\times25$.
Regarding the RPN stage, the TridentAlign module generates a feature pyramid of size $K=3$ with spatial dimensions $s_i\in\{3,5,9\}$. A total of $N=64$ ROI proposal boxes are obtained via NMS with an overlap threshold value of $0.9$. 
In the subsequent classification stage, the spatial dimension of the pooled candidate features obtained by ROIAlign was set to $s=5$. 

\smallskip

\noindent
\textbf{Training data:~}~To train the model, we used training splits of the ImageNetVID \cite{imagenet}, YouTubeBB \cite{ytbb}, GOT-10k \cite{got10k}, and LaSOT \cite{LaSOT} datasets, and pairs of query and search images are uniformly sampled from the video sequences in these datasets. 
When sampling an image pair, a video sequence was chosen randomly where the probability of choosing a certain dataset is proportional to its total size. 
For a sampled image pair, we performed random data augmentation such as horizontal flips and the addition of gaussian noise, blurring, and color jitter. 
The bounding box coordinates were also randomly augmented by $\pm1\%$ of their original width/height.

\smallskip

\noindent
\textbf{Training details:~}~We optimized the sum loss functions $L=L_{rpn}+L_{det}$, where the losses are given in Eq. \eqref{eq:loss_rpn} and Eq. \eqref{eq:loss_det} with $\lambda=1$. 
We used the Adam \cite{adam} optimizer with a batch size of four pairs to train our network. 
The learning rate was set to $10^{-4}$, and the weight decay coefficient was set to $10^{-5}$. 
For initialization, we used pretrained weights from the ResNet architectures, and during training, we freeze the weights of the residual blocks, except for the last block. 
We first trained the network for $2\times10^6$ iterations without the context embedding module and decayed the learning rate by a factor of $0.5$ halfway. 
Then, we added the context embedding module and trained the network for another $10^6$ iterations with a learning rate of $10^{-5}$. We allocated an initial burn-in phase of $10^4$ iterations, during which only the RPN was trained.
In this way, we prevent negative candidate ROIs from overwhelming the classification stage, which can stabilize the training process. 
Our model was implemented in Python using the PyTorch \cite{pytorch} library. For run-time measurements, we run and time our model on a single Nvidia RTX 2080Ti GPU.

\subsection{Quantitative Evaluation}
\label{sec:quantitative}

\textbf{Evaluation datasets and metrics:~}~We evaluated our tracker (hereafter denoted as TACT) on test sets including four large-scale visual tracking benchmark datasets: LaSOT \cite{LaSOT}, OxUvA \cite{oxuva}, TrackingNet \cite{trackingnet}, and GOT-10k \cite{got10k}. 
The parameters were fixed for all benchmarks and experiments. 
LaSOT and OxUvA are long-term tracking benchmarks whose average sequence length is longer than 1 min, whereas TrackingNet and GOT-10k have shorter sequences but include a larger number of sequences with more various classes of objects.

The \textbf{LaSOT} \cite{LaSOT} dataset is a large-scale and long-term tracking dataset consisting of $1,400$ long-term sequences with an average sequence length of $2,512$ frames (83 secs), which are annotated with the bounding box coordinates of the target object. We evaluated our tracker on the test set that includes 280 sequences under a one-pass evaluation setting, where the performance metrics are the area-under-curve (AUC) of the success plot, location precision, and normalized precision.
The \textbf{OxUvA} \cite{oxuva} dataset is used to evaluate long-term tracking performance where its development and test sets have 200 and 166 sequences, respectively, with an average length of 4,260 frames (142 secs). 
In addition to evaluating the accuracy of the predicted boxes, the tracker must also report whether the target is present/absent in a given frame. 
The performance metric is the maximum geometric mean (MaxGM) over the true positive rate (TPR) and the true negative rate (TNR).
\textbf{TrackingNet} \cite{trackingnet} is a large-scale dataset of more than $30,000$ videos collected from YouTube, of which $511$ are included in the test set. 
Similar to the other benchmarks, it uses precision, normalized precision, and the AUC of the success plot as performance metrics.
\textbf{GOT-10k} \cite{got10k} is a tracking dataset focused on the one-shot experiment setting in which the training and test sets have disjoint object classes. 
It contains $10,000$ videos of which $420$ are used as the test set. 
Trackers are evaluated by calculating the success rate (SR, with threshold values 0.5 and 0.75) and average overlap (AO) value.

\begin{table}[t]
	\caption{{Comparison on the \textbf{LaSOT} test set.} }
	\label{tab:comp_lasot}
	\centering
	\resizebox{\textwidth}{!}{
		\begin{tabular}{@{}lcccccccccccc@{}}
			\toprule
			& \begin{tabular}[c]{@{}c@{}}\textbf{TACT-18}\\ \end{tabular} & \begin{tabular}[c]{@{}c@{}}\textbf{TACT-50}\\ \end{tabular} & \begin{tabular}[c]{@{}c@{}}GlobalTrack\\ \cite{globaltrack}\end{tabular} & \begin{tabular}[c]{@{}c@{}}ATOM\\ \cite{atom}\end{tabular} & \begin{tabular}[c]{@{}c@{}}SiamRPN++\\ \cite{siamrpn++}\end{tabular} & \begin{tabular}[c]{@{}c@{}}DASiam\\ \cite{dasiamrpn}\end{tabular} & \begin{tabular}[c]{@{}c@{}}SPLT\\ \cite{splt}\end{tabular} & \begin{tabular}[c]{@{}c@{}}MDNet\\ \cite{mdnet}\end{tabular} & \begin{tabular}[c]{@{}c@{}}StructSiam\\ \cite{stsiam}\end{tabular} & \begin{tabular}[c]{@{}c@{}}SiamFC\\ \cite{siamfc}\end{tabular} & \begin{tabular}[c]{@{}c@{}}CFNet\\ \cite{cfnet}\end{tabular} & \begin{tabular}[c]{@{}c@{}}ECO\\ \cite{eco}\end{tabular} \\ \midrule
			\textbf{AUC} & \textcolor{blue}{0.556} & \textcolor{red}{0.575} & 0.521 & 0.518 & 0.496 & 0.448 & 0.426 & 0.397 & 0.335 & 0.336 & 0.275 & 0.324 \\
			\textbf{Precision} & \textcolor{blue}{0.583} & \textcolor{red}{0.607} & 0.529 & 0.506 & 0.491 & 0.427 & 0.396 & 0.373 & 0.333 & 0.339 & 0.259 & 0.301 \\
			\scriptsize{\textbf{Norm. Precision}} & \textcolor{blue}{0.638} & \textcolor{red}{0.660} & 0.599 & 0.576 & 0.569 & - & 0.494 & 0.460 & 0.418 & 0.420 & 0.312 & 0.338 \\
			\midrule
			\textbf{FPS} & 57 & 42 & 6 & 30 & 35 & 110 & 25.7 & 0.9 & 45 & 58 & 43 & 60 \\
			\bottomrule
		\end{tabular}
	}
\end{table}

\begin{table}[t]
	\caption{{Comparison on the  \textbf{OxUvA} test set.} }
	\label{tab:comp_oxuva}
	\centering
	\resizebox{\textwidth}{!}{
		\begin{tabular}{@{}lccccccccccccc@{}}
			\toprule
			(\%) & \begin{tabular}[c]{@{}c@{}}\textbf{TACT-50}\\ \end{tabular} & \begin{tabular}[c]{@{}c@{}}GlobalTrack\\ \cite{globaltrack}\end{tabular} & \begin{tabular}[c]{@{}c@{}}SPLT\\ \cite{splt}\end{tabular} &
			\begin{tabular}[c]{@{}c@{}}MBMD\\ \cite{mbmd}\end{tabular} & \begin{tabular}[c]{@{}c@{}}DASiam$_{\textrm{LT}}$\\ \cite{dasiamrpn}\end{tabular} & \begin{tabular}[c]{@{}c@{}}EBT\\ \cite{ebt}\end{tabular} & \begin{tabular}[c]{@{}c@{}}SiamFC{\tiny+R}\\ \cite{siamfc}\end{tabular} & \begin{tabular}[c]{@{}c@{}}SINT\\ \cite{sint}\end{tabular} & \begin{tabular}[c]{@{}c@{}}LCT\\ \cite{lct}\end{tabular} & \begin{tabular}[c]{@{}c@{}}TLD\\ \cite{tld}\end{tabular} &
			\begin{tabular}[c]{@{}c@{}}MDNet\\ \cite{mdnet}\end{tabular} & \begin{tabular}[c]{@{}c@{}}ECO-HC\\ \cite{eco}\end{tabular} &  \\ \midrule
			\textbf{MaxGM} & \textcolor{red}{70.9} & 60.3 & \textcolor{blue}{62.2} & 54.4 & 41.5 & 28.3 & 45.4 & 32.6 & 39.6 & 43.1 & 34.3 & 31.4\\
			\textbf{TPR} & \textcolor{red}{80.9} & \textcolor{blue}{57.4} & 49.8 & 60.9 & 68.9 & 32.1 & 42.7 & 42.6 & 29.2 & 20.8 & 47.2 & 39.5\\
			\textbf{TNR} & 62.2 & 63.3 & \textcolor{blue}{77.6} & 48.5 & 0.0 & 0.0 & 48.1 & 0.0 & 53.7 & \textcolor{red}{89.5} & 0.0 & 0.0\\
			\bottomrule
		\end{tabular}
	}
\end{table}

\smallskip

\noindent
\textbf{Comparison to other trackers:~}~We evaluated our proposed tracker on the LaSOT test set and provide the results in Table \ref{tab:comp_lasot}, where we denote our tracker with the ResNet-18 backbone as \textbf{TACT-18} and that with the ResNet-50 backbone as \textbf{TACT-50}. 
Both variants of TACT outperform other recent ResNet backbone-based trackers, which are GlobalTrack \cite{globaltrack}, ATOM \cite{atom}, SiamRPN++ \cite{siamrpn++}, and SPLT \cite{splt}. 
Moreover, our tracker runs faster than these algorithms, at real-time speed. 
To further test the long-term tracking capabilities of TACT, we evaluated our tracker on the OxUvA test set and show the results in Table \ref{tab:comp_oxuva}. 
To predict the presence/absence of the target, we simply used a threshold value of $0.95$.
Output confidence scores below the given threshold were considered to indicate absence of the target. 
The results show that our tracker outperforms other long-term tracking algorithms in terms of the MaxGM and TPR metrics by a substantial margin, even when compared to GlobalTrack \cite{globaltrack} and SPLT \cite{splt}, which are trackers specifically designed for long-term tracking applications.

\begin{table}[t]
	\caption{{Comparison on the \textbf{TrackingNet} test set.} }
	\label{tab:comp_trackingnet}
	\centering
	\resizebox{\textwidth}{!}{
	\begin{tabular}{@{}lccccccccccc@{}}
		\toprule
		(\%) & \begin{tabular}[c]{@{}c@{}}\textbf{TACT-18}\\ \end{tabular} & \begin{tabular}[c]{@{}c@{}}\textbf{TACT-50}\\ \end{tabular} & \begin{tabular}[c]{@{}c@{}}GlobalTrack\\ \cite{globaltrack}\end{tabular} & \begin{tabular}[c]{@{}c@{}}ATOM\\ \cite{atom}\end{tabular} & \begin{tabular}[c]{@{}c@{}}SiamRPN++\\ \cite{siamrpn++}\end{tabular} & \begin{tabular}[c]{@{}c@{}}DASiam\\ \cite{dasiamrpn}\end{tabular} & \begin{tabular}[c]{@{}c@{}}UPDT\\ \cite{updt}\end{tabular} & \begin{tabular}[c]{@{}c@{}}MDNet\\ \cite{mdnet}\end{tabular} & \begin{tabular}[c]{@{}c@{}}SiamFC\\ \cite{siamfc}\end{tabular} & \begin{tabular}[c]{@{}c@{}}CFNet\\ \cite{cfnet}\end{tabular} & \begin{tabular}[c]{@{}c@{}}ECO\\ \cite{eco}\end{tabular} \\ \midrule
		\textbf{Precision} & \textcolor{blue}{70.1} & \textcolor{red}{70.8} & 65.6 & 64.8 & 69.4 & 59.1 & 55.7 & 56.5 & 53.3 & 53.3 & 49.2 \\
		\scriptsize{\textbf{Norm. Precision}} & {78.4} & \textcolor{blue}{78.8} & 75.4 & 77.1 & \textcolor{red}{80.0} & 73.3 & 70.2 & 70.5 & 66.6 & 65.4 & 61.8 \\
		\textbf{Success} & \textcolor{blue}{73.4} & \textcolor{red}{74.0} & 70.4 & 70.3 & 73.3 & 63.8 & 61.1 & 60.6 & 57.1 & 57.8 & 55.4 \\ \bottomrule
	\end{tabular}
	}
\end{table}

\begin{table}[t]
	\caption{{Comparison on the \textbf{GOT-10k} test set.} }
	\label{tab:comp_got10k}
	\centering
	\resizebox{\textwidth}{!}{
	\begin{tabular}{@{}lcccccccccccc@{}}
		\toprule
		(\%) & \begin{tabular}[c]{@{}c@{}}\textbf{TACT-18}\\ \end{tabular} & \begin{tabular}[c]{@{}c@{}}\textbf{TACT-50}\\ \end{tabular} & \begin{tabular}[c]{@{}c@{}}ATOM\\ \cite{atom}\end{tabular} &
		\begin{tabular}[c]{@{}c@{}}SiamMask\\ \cite{siammask}\end{tabular} & \begin{tabular}[c]{@{}c@{}}CFNet\\ \cite{cfnet}\end{tabular} & \begin{tabular}[c]{@{}c@{}}SiamFC\\ \cite{siamfc}\end{tabular} & \begin{tabular}[c]{@{}c@{}}GOTURN\\ \cite{goturn}\end{tabular} & \begin{tabular}[c]{@{}c@{}}CCOT\\ \cite{ccot}\end{tabular} & \begin{tabular}[c]{@{}c@{}}ECO\\ \cite{eco}\end{tabular} & \begin{tabular}[c]{@{}c@{}}CF2\\ \cite{cf2}\end{tabular} & \begin{tabular}[c]{@{}c@{}}MDNet\\ \cite{mdnet}\end{tabular} &  \\ \midrule
		\textbf{$\textrm{SR}_{0.50}$} & \textcolor{blue}{64.8} & \textcolor{red}{66.5} & 63.4 & 58.7 & 40.4 & 35.3 & 37.5 & 32.8 & 30.9 & 29.7 & 30.3 \\
		\textbf{$\textrm{SR}_{0.75}$} & \textcolor{blue}{44.7} & \textcolor{red}{47.7} & 40.2 & 36.6 & 14.4 & 9.8 & 12.4 & 10.7 & 11.1 & 8.8 & 9.9 \\
		\textbf{$\textrm{AO}$} & \textcolor{blue}{55.9} & \textcolor{red}{57.8} & 55.6 & 51.4 & 37.4 & 34.8 & 34.7 & 32.5 & 31.6 & 31.5 & 29.9 \\ \bottomrule
	\end{tabular}
	}
\end{table}

We also evaluated TACT on the relatively short-term and large-scale tracking benchmarks, which are TrackingNet and GOT-10k. 
The evaluation results for TrackingNet are shown in Table \ref{tab:comp_trackingnet}, where both variants of TACT exhibit competitive results in terms of the precision and success rate metrics, outperforming most trackers. 
Furthermore, Table \ref{tab:comp_got10k} shows that the proposed method also obtains consistent results with regard to the comparison performed on GOT-10k: both variants of TACT were able to demonstrate high performance on all metrics. 
Even without any temporal smoothness assumptions or manual parameter tuning, TACT was able to achieve superior performance on the short-term tracking datasets compared to that of conventional trackers that are focused on short-term tracking applications.

\begin{figure}[t]
	\begin{center}
		\includegraphics[width=1.00\linewidth, height=0.5\linewidth]{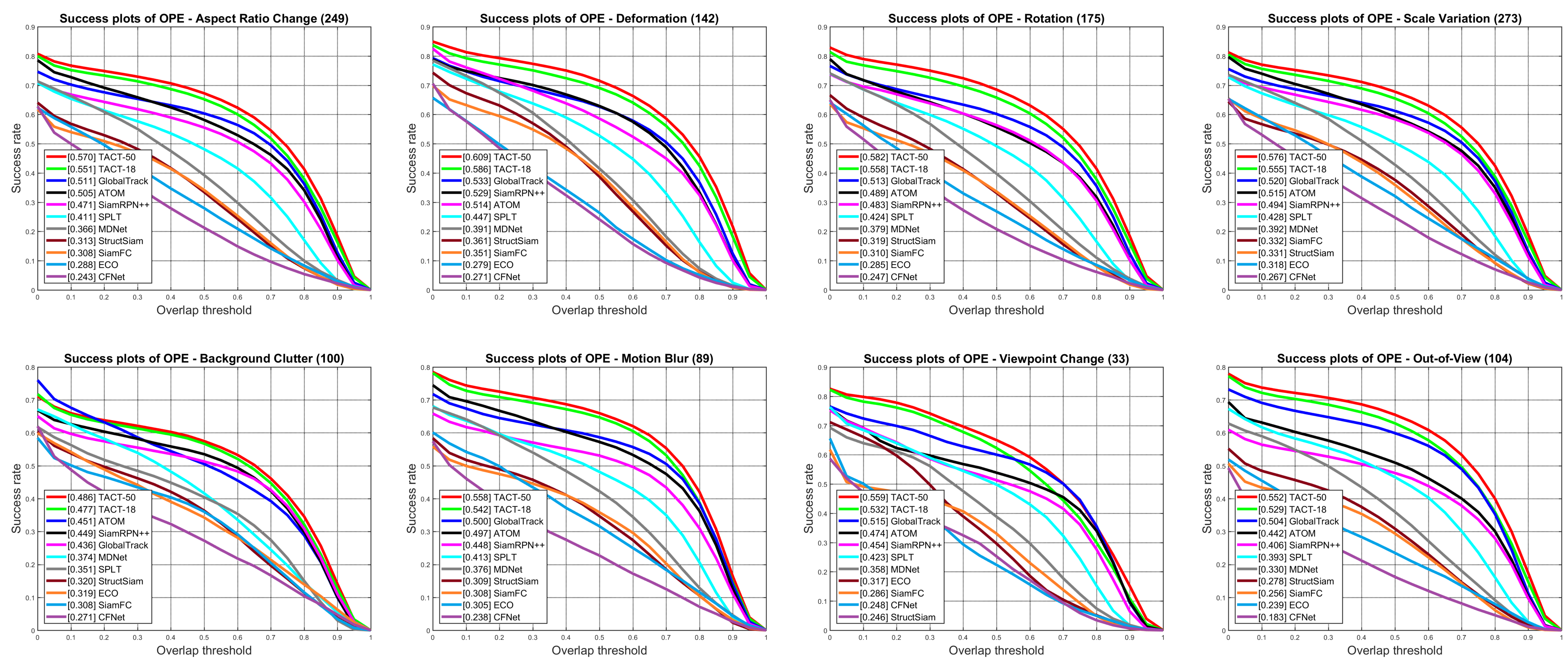}
	\end{center}
	\vspace{-1mm}
	\caption{{Success plots for eight challenge attributes of the LaSOT dataset}}
	\label{fig:attribute}
\end{figure}

To analyze the effectiveness of our proposed TridentAlign and context embedding modules, we show the success plots for eight different challenge attributes of the LaSOT dataset in Figure \ref{fig:attribute}. 
TACT achieved substantial performance improvements on the attributes of aspect ratio change, deformation, rotation, and scale variation. 
Compared to other full-frame search-based tracker GlobalTrack, TACT performed better by a large margin owing to its TridentAlign module, which facilitates robustness to scale variations and large deformations of the target. 
Moreover, GlobalTrack only considers local features without any global context information; therefore, it is more prone to being affected by distractors similar to the target as shown in the background clutter plots, with inferior performance to that of ATOM and SiamRPN++. 
Using our context embedding module, robustness to background clutter can be reinforced via the global context modeling and embedding scheme. 
TACT also shows improvements with regard to the motion blur, viewpoint change, and out-of-view attributes owing to its full-frame search based design, which allows it to successfully recover from prolonged out-of-frame target disappearances and brief drifts. 

\begin{figure}[t]
	\begin{center}
		\includegraphics[width=1.00\linewidth]{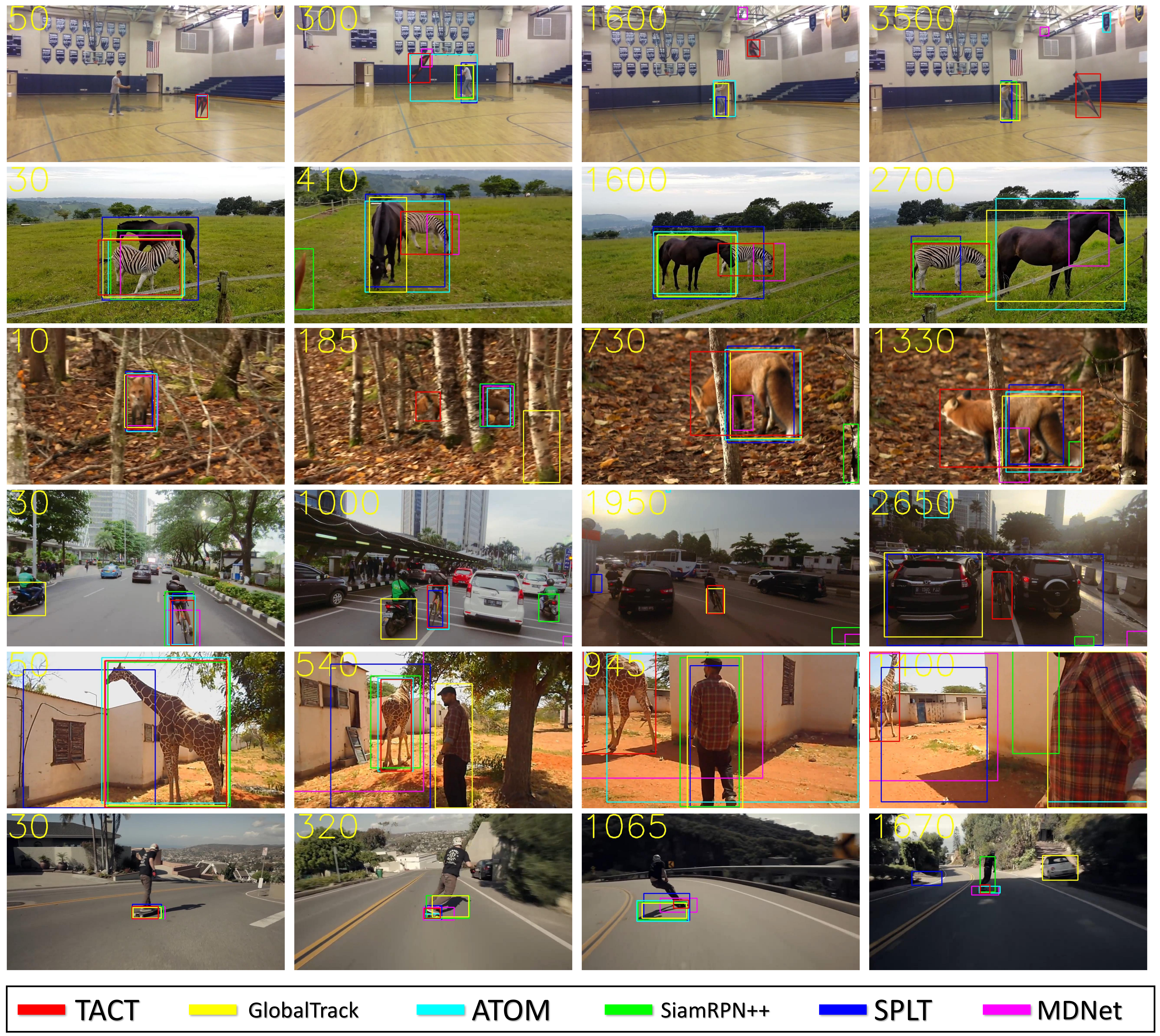}
	\end{center}
	\vspace{-7mm}
	\caption{\textbf{Qualitative Results on LaSOT.} Tracking results for the sequences \textit{kite-6}, \textit{zebra-17}, \textit{fox-5}, \textit{bicycle-9}, \textit{giraffe-10}, and \textit{skateboard-19}. The color of the bounding box denotes a specific tracker. Yellow numbers on the top-left corner indicate frame indexes.}
	\label{fig:qualitative}
		\vspace{-3mm}
\end{figure}

In Figure \ref{fig:qualitative}, we present qualitative results produced by GlobalTrack, ATOM, SiamRPN++, SPLT, MDNet, and TACT-50 for selected videos from the LaSOT dataset. 
The results show that TACT successfully tracks the target despite challenging conditions such as large deformation in \textit{kite-6}, occlusion in \textit{fox-5}, and background clutter in \textit{bicycle-9} and \textit{skateboard-19}, whereas the other trackers fail. 
For additional attribute plots and qualitative results for other videos from the LaSOT dataset, please refer to the attached supplementary document and video.

\subsection{Ablation Study}
\label{sec:ablation}

To provide more in-depth analysis of and insight into the proposed TACT, we performed additional ablation experiments for the individual components. 
For the following experiments, we used the test set of the LaSOT dataset and the success plot AUC metric to compare the performance of different variants of TACT.

\smallskip

\noindent
\textbf{Component-wise ablation:~}~To validate the contribution of each individual component to the performance gain, we compared different variants of TACT by adding or removing the proposed modules. 
Table \ref{tab:ablation_module} shows the results of the ablation analysis on individual components for both TACT-18 and TACT-50. Starting from the variants without the TridentAlign or context embedding modules, adding each component consistently improves the performance of both TACT-18 and TACT-50, which validates the effectiveness of our proposed approach. 
Our proposed TridentAlign and context embedding modules contribute to +2.1\% and +2.3\% to the performance gains of TACT-18 and TACT-50, respectively. 
All models were trained under the same settings using the same training datasets until convergence.

\smallskip

\noindent
\textbf{Variants of the context embedding module:~}~We test four possible designs for the context embedding module, as introduced in Table \ref{tab:context}. For the experiments, we started from a baseline model of TACT-18, which was trained without the context embedding module.
Then, we added the context embedding module on top of the baseline model and trained the final model for additional iterations (as specified in Section  \ref{sec:implementation}). 
Table \ref{tab:ablation_cxt} shows the results for different variants of the context embedding module. 
Among all variants, the FILM-based module shows the highest performance gain followed by the simple addition-based module and the CBAM-based module.
In contrast, the simple concatenation-based module degrades the performance. 
These results are somewhat consistent with other affine transform-based feature-wise transformation methods utilized in previous literature \cite{film}, where its multiplicative and additive operations provide adequate conditioning information for a given feature without hampering the discriminability of the original feature space.

\begin{table}[t]
	\caption{\textbf{Ablation analysis of individual components.} Adding each component contributes to consistent performance gains over the baseline model. As a performance measure, the AUC of the success plot on the LaSOT test set is shown.}
	\label{tab:ablation_module}
	\centering
	\begin{tabular}{cccc}
		\toprule
		~~ \textbf{TridentAlign} ~~ & ~~ \textbf{Context Embedding} ~~ & \begin{tabular}[c]{@{}c@{}}\textbf{AUC}\\ (TACT-18)\end{tabular} & \begin{tabular}[c]{@{}c@{}}\textbf{AUC}\\ (TACT-50)\end{tabular} \\ \midrule
		\ding{55} & \ding{55} & 0.535 & 0.552 \\
		\ding{51} & \ding{55} & 0.545 & 0.564 \\
		\ding{51} & \ding{51} & \textbf{0.556} & \textbf{0.575} \\
		\bottomrule
	\end{tabular}
\end{table}

\begin{table}[t]
	\caption{\textbf{Ablation analysis of the context embedding module.} Among possible variants of the context embedding module, the FILM-based module shows the best performance. As a performance measure, the AUC of the success plot for the LaSOT test set is shown.}
	\label{tab:ablation_cxt}
	\centering
	\resizebox{\textwidth}{!}{
		\begin{tabular}{cccccc}
			\toprule
			&~~\textbf{No context}~~&~~\textbf{Simple concat.}~~&~~\textbf{Simple add.}~~& ~~\textbf{CBAM-based}~~&~~\textbf{FILM-based}~~\\ \midrule
			\begin{tabular}[c]{@{}c@{}}\textbf{AUC}\\ (TACT-18)\end{tabular} & 0.545 &0.532 & 0.552 & 0.551 & \textbf{0.556} \\
			\bottomrule
		\end{tabular}
	}
\end{table}

\pagebreak

\section{Conclusion}
\label{sec:conclusion}
	\vspace{-3mm}
In this paper, we proposed a novel visual tracking method that aims to deal with large scale variations and deformations of the target while improving its discriminability by utilizing the global context information of the surroundings. 
Built upon a two-stage object detection framework, our proposed TACT incorporates the TridentAlign and context embedding modules to overcome the limitations of conventional tracking algorithms. 
The TridentAlign module constructs a target feature pyramid that encourages the adaptability of the tracker to large scale variations and deformations by fully utilizing the spatial information of the target. 
The context embedding module generates and embeds the global context information of a given frame into a local feature representation for improved discriminability of the target against distractor objects. 
The proposed modules are designed efficiently such that the overall framework can run at a real-time speed. 
Experimental results on four large-scale visual tracking benchmarks validate the strong performance of the TACT on long-term and short-term tracking tasks, achieving improved performance on challenging scenarios.

\bibliographystyle{splncs}

\end{document}